\documentclass[runningheads]{llncs}
\usepackage[T1]{fontenc}
\usepackage{graphicx}
\usepackage{multicol}
\usepackage{multirow}
\usepackage{amsmath,amssymb,amsfonts}%
\usepackage{mathrsfs}%
\usepackage[title]{appendix}%
\usepackage{xcolor}%
\usepackage{textcomp}%
\usepackage{booktabs}%
\usepackage{manyfoot}%
\usepackage{algorithm}%
\usepackage{algorithmicx}%
\usepackage{algpseudocode}%
\usepackage{listings}%
\usepackage{pgfplots} 
\pgfplotsset{compat=1.15}
\usepackage{mathrsfs}
\usepackage{makecell}
\usepackage{subcaption}
\usepackage{xspace}
\usepackage[table]{xcolor}
\usepackage{hyperref}
\usepackage{tabularx}
\usepackage{array}

\newcolumntype{L}{>{\raggedright\arraybackslash}X}

\newcommand{\etal}{\textit{et al. }}

\newcommand{\up}{$\textcolor{red}{\uparrow}$}
\newcommand{\down}{$\textcolor{blue}{\downarrow}$}

\definecolor{zzttqq}{rgb}{0.6,0.2,0}
\definecolor{ududff}{rgb}{0.3,0.3,1}

\definecolor{pastelgreen}{RGB}{190,225,200}
\definecolor{pastelred}{RGB}{245,195,190}

\begin{document}
\title{HANCLIP: Lightweight Geometric Fine-Tuning for Negation-Aware Multimodal Search}
\titlerunning{HANCLIP}
%
\author{Hoang-Bao Le\inst{1} \and
Aiden Durrant\inst{2} \and
Thai Son Mai\inst{3} \and
Binh T. Nguyen\inst{4} \and
Liting Zhou\inst{1} \and
Cathal Gurrin\inst{1}}
\authorrunning{H.-B. Le et al.}
\institute{ADAPT Centre, Dublin City University, Dublin, Ireland \and
University of East Anglia, Norwich, UK \and
Queen's University Belfast, Belfast, UK \and
University of Science, Vietnam National University, Ho Chi Minh City, Vietnam}
\maketitle              
\begin{abstract}
Vision-language models (VLMs) achieve strong cross-modal alignment but remain brittle to negation, often relying on shallow word associations rather than compositional reasoning. Fine-tuning on negation-specific data can also compromise their general purpose capabilities through catastrophic forgetting. We introduce \textbf{HANCLIP} (\textbf{H}yperbolic, \textbf{A}ngular, and \textbf{N}egation), a geometry-aware framework that improves negation sensitivity while preserving the structure of the pretrained joint embedding space. HANCLIP combines a hyperbolic contrastive objective, which models hierarchical relations and semantic asymmetries, with an angular triplet loss that separates negated descriptions from their affirmative counterparts. Using only 20,000 image-text quadruplets, HANCLIP consistently improves performance across CLIP, LongCLIP, and SmartCLIP backbones on the NegBench benchmark, while maintaining or improving zero-shot classification and image-text retrieval performance. These results show that lightweight, geometry-guided objectives can enhance negation understanding without large-scale retraining.

\end{abstract}

\section{Introduction}


Negation is a fundamental logical operation that maps a proposition $P$ to its complement, \emph{not} $P$. While linguistic negation is often expressed through simple syntactic transformations, grounding negation in visual context is substantially more complex. Despite impressive progress in vision–language modeling, recent studies reveal that Vision–Language Models (VLMs) struggle with such negation understanding \cite{alhamoud2025negbench,park2025negationclip}. Yuksekgönül \etal \cite{yuksekgonul2023negclip} demonstrate that models trained with standard contrastive objectives primarily learn shallow word–object associations, aligning individual nouns with visual entities while failing to capture relational and logical structure. As a result, VLMs often assign high similarity scores to both affirmative and negated captions, leading to unreliable behaviour in real-world applications that demand precise semantic understanding.

\begin{figure}[!t]
    \centering
    \includegraphics[width=.8\linewidth]{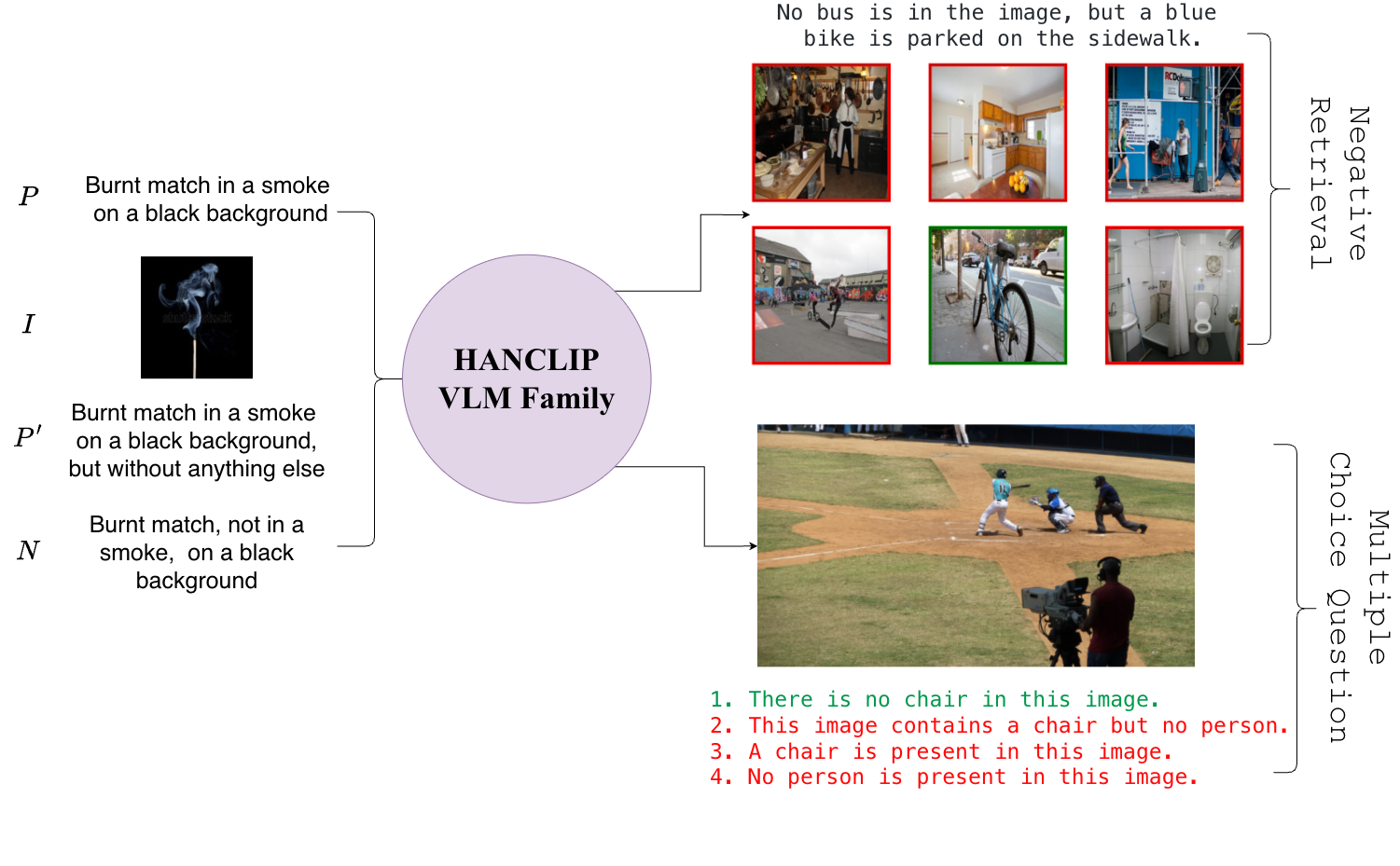}
    \caption{Overview of HANCLIP Framework with a quadruplet input of $(I,P,P',N)$ and two examples on NegBench.}
    \label{fig:hanclip}\vspace{-2em}
\end{figure}

To address these limitations, a growing body of work proposes negation-aware datasets, specialised contrastive objectives \cite{park2025negationclip,yuksekgonul2023negclip}, and challenging NegBench \cite{alhamoud2025negbench} benchmark. While effective, these approaches typically rely on large-scale curated data and extensive retraining, which limits their practicality and reproducibility. Furthermore, while improving negation sensitivity is a priority, it often leads to knowledge forgetting \cite{le2024ceclip,yuksekgonul2023negclip}. This creates a trade-off where gains in negation handling can reduce the performance of VLMs on standard retrieval and zero-shot tasks. 

Recently, hyperbolic representation learning has emerged as a promising alternative to Euclidean embedding spaces for modeling hierarchical and asymmetric\cite{ermolov2022hyperbolic}. Several hyperbolic VLMs have been proposed, including MERU \cite{desai2023meru}, HyCoCLIP \cite{pal2024hycoclip}, and HyperVLM \cite{srivastava2025hypervlm}. While MERU \cite{desai2023meru} emphasises inter-modal partial ordering between images and texts, but overlooks intra-modal semantic structure, HyCoCLIP \cite{pal2024hycoclip} addresses this limitation by modeling object-level hierarchies through region-based alignments. HyperVLM \cite{srivastava2025hypervlm} further generalises partial ordering by dynamically inferring modality dominance on a per-instance basis. These works suggest that hyperbolic geometry is well-suited for capturing semantic hierarchies, yet its potential for negation modeling remains underexplored \cite{yoshikawa2025phyclip,pal2024hycoclip}. 

In this study, to address the aforementioned challenges of negation understanding in VLMs, we introduce \underline{H}yperbolic \underline{A}ngular \underline{N}egation CLIP (HANCLIP), a family of VLMs equipped with two complementary components: (i) Hyperbolic Contrastive Objective (HCO) and (ii) Angular Triplet Negation Loss (ATNL).
Instead of relying solely on Euclidean pushing, the Hyperbolic Contrastive Objective (HCO) uses hyperbolic distance over inter-textual negatives to structure a hierarchy distinguishing affirmative concepts from their negated counterparts. This forces affirmative captions and negated-positive pairs to remain semantically close while isolating incompatible negatives in separate hyperbolic regions. To complement HCO, the Angular Triplet Negation Loss (ATNL) introduces geometry-aware regularisation operating on relative directions rather than absolute distances. Specifically, ATNL enforces angular alignment between negated-positive and affirmative captions relative to a negative anchor while maximising separation between opposing semantic directions. By optimising these cosine-based angular relationships, ATNL prevents spurious lexical correlations and stabilises global image-text alignment. Finally, we demonstrate the effectiveness of our approach through experiments on three VLM backbones: CLIP \cite{radford2021learning}, LongCLIP \cite{zhang2024longclip}, and SmartCLIP \cite{xie2025smartclip} across diverse benchmarks.

In summary, we make the following contributions:

\vspace{-.5em}
\begin{itemize}
    \item Firstly, we use Hyperbolic space to solve the Negation Understanding knowledge of Vision Language Models. Compared to the original VLMs, by leveraging an non-Euclidean metric space, our HANCLIP VLM Family delivers better understanding on negation-containing problems. 
    \item Secondly, we introduce Angular Triplet Negation Loss to pull the True positive sentence and push the Negative one away the anchor embedding. 
    \item Thirdly, with 20,000 training samples, extensive experiments across CLIP, LongCLIP, and SmartCLIP on several benchmarks show that our approach improve their performance with an average improvement of more than 17\% in Negative Retrieval and nearly 4.4\% in Multiple Choice Question tasks. 
\end{itemize}

\vspace{-1.5em}\section{Methodology}
\subsection{Preliminaries}

Let $V \in \mathcal{V}$ denote an image paired with three types of captions: a positive caption $P \in \mathcal{P}$, a negated-positive caption $P' \in \mathcal{P}'$, and a negative caption $N \in \mathcal{N}$. We consider a pre-trained Vision-Language Model (VLM) $f(\cdot)$ that encodes both images and texts into a shared $d$-dimensional embedding space. The corresponding representations are denoted by $\mathbf{v}$, $\mathbf{p}$, $\mathbf{p}'$, and $\mathbf{n} \in \mathbb{R}^d$, respectively.

\begin{figure}[!ht]
    \centering
    \includegraphics[width=0.7\linewidth]{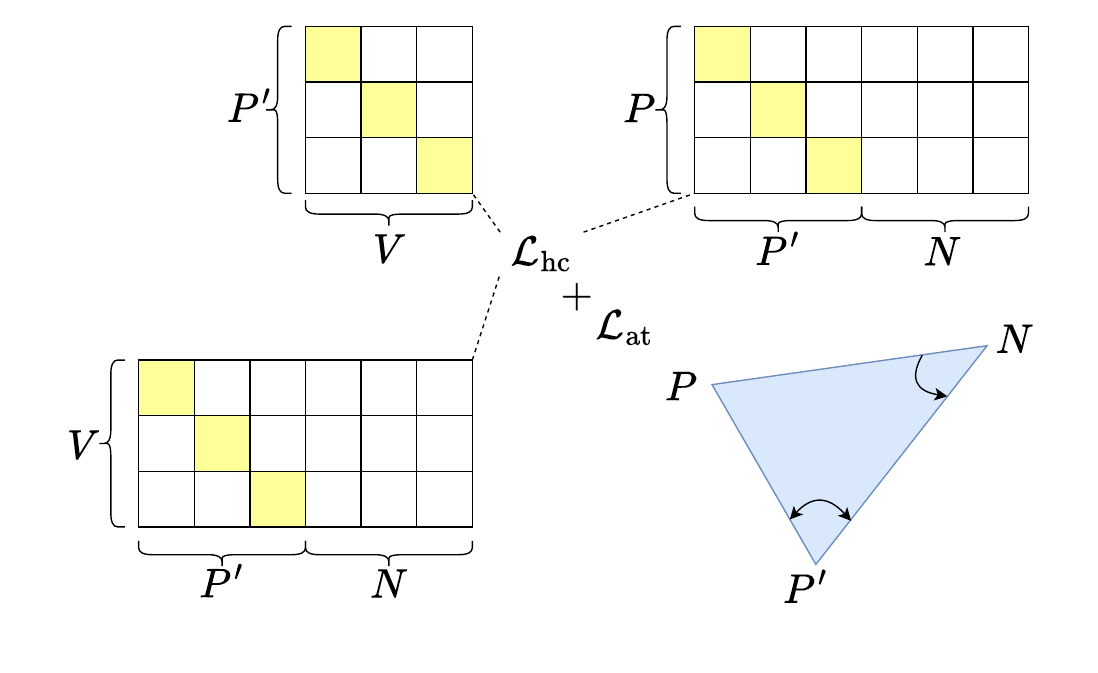}
    \caption{We incorporate negation samples in Hyperbolic Contrastive Objective $\mathcal{L}_\text{hc}$ and Angular Triplet Negation Loss $\mathcal{L}_\text{at}$ to help pre-trained VLM distinguish ground-truth captions from opposite groups.} 
    \label{fig:loss}
\end{figure}

\subsection{Hyperbolic Geometry Background}

Hyperbolic space $\mathbb{H}^n$ is a non-Euclidean geometry with constant negative curvature, which naturally models data with hierarchical or tree-like structure. Compared to Euclidean space, distances in hyperbolic space grow exponentially with radius, allowing it to represent inclusion relations and semantic hierarchies more efficiently. This property has recently motivated its use in representation learning and vision-language models.
We formulate our method in the \emph{Poincar\'e ball model} of hyperbolic geometry~\cite{ermolov2022hyperbolic}, chosen for its numerical stability and compatibility with gradient‑based optimisation. A Poincaré ball is an open $n$-dimensional ball
\begin{equation}
\mathbb{D}_c^n = \{ \mathbf{x} \in \mathbb{R}^n \mid c\|\mathbf{x}\|_2^2 < 1 \},
\end{equation}
where $c \ge 0$ controls the curvature (with actual curvature $-c^2$). 

Since hyperbolic space is not a vector space, standard Euclidean operations such as addition do not apply. Instead, we use the \emph{gyrovector} formalism , which provides algebraic operations consistent with the underlying geometry. For $\mathbf{x}, \mathbf{y} \in \mathbb{D}_c^n$, the Möbius addition is defined as
\begin{equation}
\mathbf{x} \oplus_c \mathbf{y}
=
\frac{
(1 + 2c\langle \mathbf{x}, \mathbf{y} \rangle + c\|\mathbf{y}\|_2^2)\mathbf{x}
+ (1 - c\|\mathbf{x}\|_2^2)\mathbf{y}
}{
1 + 2c\langle \mathbf{x}, \mathbf{y} \rangle + c^2\|\mathbf{x}\|_2^2\|\mathbf{y}\|_2^2
}.
\end{equation}
The hyperbolic distance between two points $\mathbf{x}, \mathbf{y} \in \mathbb{D}_c^n$ is given by
\begin{equation}
D_h(\mathbf{x}, \mathbf{y})
=
\frac{2}{\sqrt{c}}
\tanh^{-1}
\!\left(
\sqrt{c}\,
\| -\mathbf{x} \oplus_c \mathbf{y} \|_2
\right),
\end{equation}
where $\|\cdot\|_2$ denotes the Euclidean norm.
To embed Euclidean features into hyperbolic space, given a base point $\mathbf{x} \in \mathbb{D}_c^n$ and a tangent vector $\mathbf{a} \in \mathbb{R}^n$, we employ the Riemannian exponential map as
\begin{equation}
\exp_\mathbf{x}^c(\mathbf{a})
=
\mathbf{x}
\oplus_c
\left(
\tanh\!\left(
\frac{\sqrt{c}\lambda_\mathbf{x}^c \|\mathbf{a}\|_2}{2}
\right)
\frac{\mathbf{a}}{\sqrt{c}\|\mathbf{a}\|_2}
\right).
\end{equation}
In practice, we use the origin as the base point, which simplifies the mapping and ensures stable optimisation.
\vspace{-.5em}
\subsection{Hyperbolic Contrastive Objective}
As illustrated in Figure~\ref{fig:loss}, our proposed HCO explicitly model negation using hyperbolic geometry and is guided by a key observation: while $P$ and $P'$ may differ syntactically, they describe closely related visual contents, whereas $N$ describes a genuinely incompatible concept. In hyperbolic geometry, semantically related concepts are embedded closer to each other near the centre of the space, while unrelated concepts are pushed toward the boundary. We therefore embed image and text features into the Poincar\'e ball, where hierarchical separation between $(P, P')$ and $N$ can be expressed through hyperbolic distance.

Concretely, let $g(\cdot)=\exp_0^c(\cdot)$ denote the exponential map from Euclidean space to the hyperbolic manifold with curvature $c$. Unlike the standard CLIP objective, which treats all non-matching samples as implicit negatives, our HCO explicitly includes the true negative captions $\mathcal{N}$ in the denominator of the contrastive loss. This design enforces stronger discrimination between negated positives and genuine negatives. Given a batch of size $B$, temperature $\tau$, a set of image features $\mathcal{V}$, and the corresponding triplet captions $(\mathcal{P}, \mathcal{P}', \mathcal{N})$, the proposed HCO loss $\mathcal{L}_{\mathrm{hc}}$ consists of three complementary terms $\mathcal{L}_{\mathcal{P}' \to \mathcal{V}}$, $\mathcal{L}_{\mathcal{V} \to \mathcal{P}' \cup \mathcal{N}}$ and $\mathcal{L}_{\mathcal{P}' \to \mathcal{P} \cup \mathcal{N}}$ as follows:
\begin{equation}
\mathcal{L}_{\mathcal{P}' \to \mathcal{V}} = -\frac{1}{|B|}\sum_{i\in B}\log \frac{\exp\!\left(-\texttt{dist}(\mathbf{p}'_i, \mathbf{v}_i)/\tau\right)}{\sum_{k=1}^B \exp\!\left(-\texttt{dist}(\mathbf{p}'_i, \mathbf{v}_k)/\tau\right)}
\end{equation}
\begin{equation}
\mathcal{L}_{\mathcal{V} \to \mathcal{P}' \cup \mathcal{N}} = -\frac{1}{|B|} \sum_{i \in B} \log \frac{\exp\!\left(-\texttt{dist}(\mathbf{v}_i, \mathbf{p}'_i)/\tau\right)}{\begin{aligned}\sum_{k=1}^B \left[ \exp\!\left(-\texttt{dist}(\mathbf{v}_i, \mathbf{p}'_k)/\tau\right) + \exp\!\left(-\texttt{dist}(\mathbf{v}_i, \mathbf{n}_k)/\tau\right) \right]\end{aligned}}
\end{equation}
\begin{equation}
\mathcal{L}_{\mathcal{P}' \to \mathcal{P} \cup \mathcal{N}} = -\frac{1}{|B|} \sum_{i \in B} \log \frac{\exp\!\left(-\texttt{dist}(\mathbf{p}'_i, \mathbf{p}_i)/\tau\right)}{\begin{aligned}\sum_{k=1}^B \left[ \exp\!\left(-\texttt{dist}(\mathbf{p}'_i, \mathbf{p}_k)/\tau\right) + \exp\!\left(-\texttt{dist}(\mathbf{p}'_i, \mathbf{n}_k)/\tau\right) \right]\end{aligned}}
\end{equation}
\begin{equation}\label{l_hc}
    \mathcal{L}_{hc} = \mathcal{L}_{\mathcal{P}' \to \mathcal{V}}+ \mathcal{L}_{\mathcal{V} \to \mathcal{P}' \cup \mathcal{N}} + \mathcal{L}_{\mathcal{P}' \to \mathcal{P} \cup \mathcal{N}}
\end{equation}
where $\texttt{dist}(\textbf{x}, \textbf{y})=D_h(g(\textbf{x}), g(\textbf{y}))$,  $\mathcal{L}_{\mathcal{P}' \to \mathcal{V}}$ encourages each negated-positive caption $P'_i$ to be aligned with its corresponding image $I_i$ while contrasting against other images in the batch, $\mathcal{L}_{\mathcal{V} \!\to\! \mathcal{P}' \cup \mathcal{N}}$ enforces that an image $I_i$ is closer to its negated-positive caption $P'_i$ than to any true negative caption $N_k$, explicitly including genuine negatives in the denominator to strengthen discrimination, and $\mathcal{L}_{\mathcal{P}' \!\to\! \mathcal{P} \cup \mathcal{N}}$ aligns negated-positive captions with their corresponding positive captions $P_i$ while simultaneously separating them from negative captions $N_k$, ensuring semantic consistency between affirmative and negated descriptions.

\subsection{Angular Triplet Negation Loss in Hyperbolic Setting}

As illustrated in Figure~\ref{fig:loss}, ATNL explicitly models negation reasoning by constraining the relative \emph{angular geometry} among a positive caption $\mathbf{p}$, its negated-positive counterpart $\mathbf{p'}$, and a negative caption $\mathbf{n}$. Accordingly, the Angular Triplet Negation Loss $\mathcal{L}_{\mathrm{at}}$ consists of two complementary components: (i) a \emph{positive-alignment} term $\mathcal{L}_{\mathrm{pos}}$, which minimizes the angle between $\overrightarrow{NP}$ and $\overrightarrow{NP'}$, and (ii) a \emph{negation-separation} term $\mathcal{L}_{\mathrm{neg}}$, which maximizes the angle between $\overrightarrow{P'P}$ and $\overrightarrow{P'N}$. Together, these constraints ensure that negation-aware captions preserve semantic alignment with their positives while remaining discriminative from genuine negatives. 



The positive-alignment loss from the negative caption $\mathbf{n}$ is defined as
\begin{equation}\vspace{-.5em}
    \mathcal{L}_{\text{pos}} = \mathbb{E}[1 - \cos\theta_{\text{pos}}] = \mathbb{E}[1-\cos(\overrightarrow{NP}, \overrightarrow{NP'})]
\end{equation}
which is minimised when $\cos\theta_{\text{pos}}$ approaches one. This encourages the negated caption $\mathbf{p'}$ to lie in a similar direction to the true positive $\mathbf{p}$ when viewed from the negative anchor $\mathbf{n}$, effectively pulling the two positive variants closer together in angular space.

To enforce separation from the negative caption, we next consider directions originating from the positive caption $\mathbf{p}$ in the negative-separation loss as
\begin{equation}\label{l_hang}
    \mathcal{L}_{\text{neg}} = \mathbb{E}[\cos\theta_{\text{neg}}]=\mathbb{E}[\cos(\overrightarrow{P'N}, \overrightarrow{P'P})],
\end{equation}
which is minimised to encourage a large angular difference between $\overrightarrow{P'N}$ and $\overrightarrow{P'P}$. This pushes the negated caption away from the negative caption along distinct semantic directions.

The overall Angular Triplet Negation Loss is given by
\begin{equation}\label{l_ang}
\mathcal{L}_{\text{at}} = \mathcal{L}_{\text{pos}} + \mathcal{L}_{\text{neg}} = \mathbb{E}[1 - \cos\theta_{\text{pos}}] + \mathbb{E}[\cos\theta_{\text{neg}}].
\end{equation}
\noindent\textbf{HANCLIP training objective} In summary, the training objective of HANCLIP is a weighted sum of loss terms in Eq. \ref{l_hc} and \ref{l_hang}:
\begin{equation}\label{final_loss}
    \mathcal{L}_\text{final} = \mathcal{L}_\text{hc} + \alpha \mathcal{L}_\text{at}.
\end{equation} 
Here, $\alpha$ is a balance hyperparameter between HCO and ATNL.

\section{Experiments}

\subsection{Implementation Details}

All the images are resized to $224\times 224$. We use a curvature with $c=0.1$ for the exponential map. The temperature $\tau=0.01$ for all experiments. For the loss function in Eq. \ref{final_loss}, we set $\alpha =1$ for all experiments. All experiments are implemented on 1 GPU A100 NVIDIA-80G with $1$ epoch, weight decay $5\texttt{e}-2$, warm-up iterations $200$, batch size $64$ and random seed $0$. The model is optimised by AdamW \cite{loshchilov2017adamw}. We follow the settings at CLIP-LoRA \cite{zanella2024loraclip} for LoRA Adaptation in our code. Both the visual and textual encoder are fine-tuned using low-rank decomposition \cite{hu2022lora} with low-rank factor $r = 16, \alpha_{LoRA}=4$ on $Q,K,V$ projection layers. Especially, we set two learning rate $5\texttt{e}-6$ and $1\texttt{e}-4$ in all experiments with and without LoRA respectively. 

We randomly select 20,000 samples from CC-Neg \cite{singh2025conclip} for training. To create the negated-positive group $\mathcal{P}'$, we utilise one in twenty six special templates in Table \ref{template}. In this work, we apply our approach on three models divided into two groups: CLIP\footnote{\hyperlink{https://github.com/openai/CLIP}{https://github.com/openai/CLIP}} \cite{radford2021learning} (B/32, B/16 and L/14), LongCLIP \cite{zhang2024longclip} and SmartCLIP \cite{xie2025smartclip}. While CLIP is popularly applied on downstream tasks, the other two are designed upon ViT-B/16 and L/14 architecture to handle longer captions with 248 tokens for text encoder. 
\begin{table}[t]
\centering
\label{tab:negation_prompts}
\small
\begin{tabularx}{\linewidth}{@{}p{0.28\linewidth}L@{}}
\toprule
\textbf{Negation Type} & \textbf{Prompt Variants} \\
\midrule

\textbf{Double negation} &
It is not true that not \texttt{\{caption\}}. \newline
It is not the case that \texttt{\{caption\}} is not happening. \newline
It is not false that \texttt{\{caption\}}.
\\[3pt]

\textbf{Implicit affirmation via negation} &
There is no doubt that \texttt{\{caption\}}. \newline
One cannot deny that \texttt{\{caption\}}. \newline
It cannot be denied that \texttt{\{caption\}}. \newline
It is not in doubt that \texttt{\{caption\}}. \newline
There is no uncertainty that \texttt{\{caption\}}.
\\[3pt]

\textbf{Negation of alternatives} &
It is not something else; rather, \texttt{\{caption\}}. \newline
It is not the case that something else is happening; instead, \texttt{\{caption\}}. \newline
Not something else, but \texttt{\{caption\}}.
\\[3pt]

\textbf{Evidence-based negation} &
No evidence suggests that not \texttt{\{caption\}}. \newline
There is nothing indicating that not \texttt{\{caption\}}. \newline
There is no reason to believe that not \texttt{\{caption\}}. \newline
Nothing contradicts the fact that \texttt{\{caption\}}. \newline
No part of the scene suggests that not \texttt{\{caption\}}. \newline
There is no indication against \texttt{\{caption\}}.
\\[3pt]

\textbf{Uncertainty framing} &
We cannot say that \texttt{\{caption\}} is not true. \newline
It is not impossible that \texttt{\{caption\}}.
\\[3pt]

\textbf{Contradiction-free statements} &
Without any contradiction, \texttt{\{caption\}}. \newline
\texttt{\{caption\}}, without anything suggesting otherwise.
\\[3pt]

\textbf{Rhetorical forms} &
Why would it not be true that \texttt{\{caption\}}? \newline
Who would say that \texttt{\{caption\}} is not happening?
\\[3pt]

\textbf{Other negation forms} &
It is not true that the opposite of \texttt{\{caption\}} holds. \newline
It is not the case that \texttt{\{caption\}} never happens. \newline
\texttt{\{caption\}} is not absent.
\\

\bottomrule
\end{tabularx}
\caption{Negation-based prompt variants used for constructing the evaluation set. 
\texttt{\{caption\}} denotes the original image caption.}\label{template}\vspace{-2em}
\end{table}

\subsection{Negation Understanding}
\textbf{Datasets and Metrics.} We select \textbf{\texttt{NegBench}}\cite{alhamoud2025negbench} containing two tasks: Negative Retrieval and Multiple Choice Question (MCQ). It focuses on retrieval with negation and multiple-choice questions with negated captions, revealing that many modern models perform near chance level on negation-sensitive queries.  
In the Negative Retrieval task, the caption is modified by adding negation factors. Furthermore, each MCQ question comprises one image following by one correct answer and three wrong answers to trick the model if it misunderstands. Their format can be described as:
\begin{itemize}\vspace{-.5em}
    \item \textbf{Affirmed}: ``This image includes A (and C).''
    \item \textbf{Negated}: ``This image does not include B.''
    \item \textbf{Hybrid}: ``This image includes A but not B.''
\end{itemize}\vspace{-.5em}
We utilise Recall at K (R@K) with $K\in\{1,5,10\}$ on Negative Retrieval task, and Accuracy by the query template on Zero-shot MCQ-Neg Task.

\begin{table}[!ht]
\centering
\resizebox{\linewidth}{!}{
\begin{tabular}{l l l | ccc ccc c | ccc c}
\toprule
& & & \multicolumn{7}{c|}{\textbf{Retrieval-Neg Task}}
& \multicolumn{4}{c}{\textbf{Zero-shot MCQ-Neg Task}} \\
\cmidrule(lr){4-10} \cmidrule(lr){11-14}

\textbf{Model} & \textbf{ViT} & \textbf{Setting}
& \multicolumn{3}{c}{Image-to-Text}
& \multicolumn{3}{c}{Text-to-Image}
& \multirow{2}{*}{\textbf{rSum}}
& \multicolumn{4}{c}{MCQ Type}
\\
\cmidrule(lr){4-6}\cmidrule(lr){7-9} \cmidrule(lr){11-14}& & 
& R@1 & R@5 & R@10
& R@1 & R@5 & R@10
& 
& Affirmed & Negated & Hybrid
& \textbf{AVG} \\
\midrule

\multirow{2}{*}{NegCLIP\cite{yuksekgonul2023negclip}}
& \multirow{2}{*}{B/32} & Base   & 56.04 & 80.52 & 88.18 & 39.88 & 67.17 & 77.34 & 409.14 & 47.24 & 14.76 & 16.40 & 26.48 \\
& & + Ours & 57.00 & 81.46 & 88.98 & 41.22 & 68.73 & 78.50 & 415.89 & 62.55 & 15.50 & 28.78 & 36.01 \\
\midrule

\multirow{2}{*}{NegCLIP$_\text{CC12M}$\cite{alhamoud2025negbench}}
& B/32 & Base   & 58.34 & 82.04 & 88.96 & 41.18 & 68.10 & 78.26 & 416.88 & 80.22 & 27.86 & 60.09 & 56.81 \\
 & & + Ours & 58.26 & 81.80 & 89.46 & 42.03 & 69.14 & 79.13 & 419.82 
 & \textbf{81.89} & 28.00 & 60.50 & 57.37 \\
\midrule

\multirow{2}{*}{CLIP$_\text{CC12M}$\cite{alhamoud2025negbench}}
& B/32 & Base   & 48.72 & 73.24 & 81.82 & 30.84 & 55.81 & 66.63 & 357.06 & 72.98 & \textbf{34.06} & 56.41 & 55.04 \\
& & + Ours & 52.46 & 75.82 & 84.42 & 33.48 & 58.71 & 69.51 & 374.41 & 81.45 & 31.34 & \textbf{60.59} & \textbf{58.51} \\
\midrule
\multirow{6}{*}{CLIP\cite{radford2021learning}}
& \multirow{2}{*}{B/16}
& Base    & 47.86 & 72.36 & 81.46 & 27.05 & 50.26 & 61.31 & 340.30 & 73.03 & 10.53 & 39.36 & 41.82 \\
& & + Ours & 51.66 & 76.60 & 84.62 & 33.45 & 58.43 & 68.97 & 373.72 & 70.72 & 14.39 & 40.36 & 42.58 \\
\cline{2-14}

& \multirow{2}{*}{B/32}
& Base  & 45.34 & 69.72 & 79.08 & 24.93 & 47.93 & 59.41 & 326.42 & 69.24 & 6.90 & 39.46 & 39.40 \\
& &  + Ours & 50.30 & 74.94 & 83.14 & 30.59 & 56.02 & 67.17 & 362.15 & 62.40 & 10.86 & 43.39 & 39.63 \\
\cline{2-14}

& \multirow{2}{*}{L/14}
& Base    & 50.88 & 74.26 & 82.42 & 29.06 & 52.38 & 62.95 & 351.96 & 62.40 & 7.65 & 41.55 & 37.99 \\
&    &  + Ours & 53.12 & 76.64 & 84.96 & 34.60 & 59.10 & 69.44 & 377.86 & 66.83 & 8.88 & 42.00 & 40.06 \\
\midrule

\multirow{4}{*}{LongCLIP \cite{zhang2024longclip}}
& \multirow{2}{*}{B/16} & Base   & 52.08 & 76.28 & 83.96 & 35.83 & 61.37 & 71.70 & 381.22 & 65.06 & 12.30 & 26.69 & 35.32 \\
&   & + Ours & 53.84 & 77.66 & 85.68 & 38.18 & 63.94 & 74.13 & 393.43 & 71.85 & 15.45 & 33.00 & 40.80 \\
\cline{2-14}

& \multirow{2}{*}{L/14} & Base   & 56.36 & 79.28 & 86.80 & 40.95 & 66.10 & 75.67 & 405.15 & 64.67 & 11.44 & 33.70 & 37.30 \\
&   & + Ours & 61.70 & 84.04 & 90.26 & 44.66 & 69.45 & 78.56 & 428.66 & 72.79 & 11.55 & 40.85 & 42.56 \\
\midrule

\multirow{4}{*}{SmartCLIP \cite{xie2025smartclip}}
& \multirow{2}{*}{B/16} & Base   & 57.44 & 80.42 & 87.82 & 38.48 & 64.56 & 74.44 & 403.17 & 51.08 & 15.88 & 17.05 & 28.37 \\
&   & + Ours & 58.00 & 81.22 & 88.56 & 41.25 & 67.45 & 76.89 & 413.37 & 65.90 & 18.45 & 29.82 & 38.62 \\
\cline{2-14}

& \multirow{2}{*}{L/14} & Base   & 63.00 & 84.00 & 90.02 & 43.75 & 68.77 & 77.96 & 427.49 & 61.27 & 13.32 & 32.26 & 36.24 \\

&   & + Ours & \textbf{64.28} & \textbf{84.92} & \textbf{91.28} & \textbf{46.29} & \textbf{71.44} & \textbf{80.15} & \textbf{438.36} & 69.39 & 14.65 & 40.16 & 42.58 \\
\bottomrule
\end{tabular}}
\vspace{0.6em}

\caption{Results on the Retrieval-Neg and Zero-shot MCQ-Neg tasks using different models and ViT backbones from the NegBench evaluation. In each column, \textbf{bold} values indicate the best overall performance.}
\label{tab:negbench}\vspace{-2em}
\end{table}

\noindent\textbf{Baselines.} \label{negbench_baseline} We compare our HANCLIP family against NegationCLIP \cite{park2025negationclip}, NegCLIP \cite{yuksekgonul2023negclip} and two versions of NegCLIP and CLIP pre-trained on Negated CC12M \cite{alhamoud2025negbench}. Moreover, we retrain the Negative Understanding models under our settings to observe the changes in Table \ref{tab:negbench}.



\noindent\textbf{Discussion.} Table~\ref{tab:negbench} shows that integrating our proposed method (\textit{+ Ours}) yields consistent performance improvements across diverse vision-language backbones on NegBench. On Retrieval-Neg, overall retrieval scores ($rSum$) improve across all architectures, with $\text{SmartCLIP}_{\text{L/14}}$ achieving the benchmark's peak performance boosting I2T $R@1$ from 63.00 to 64.28, T2I $R@1$ from 43.75 to 46.29, and $rSum$ from 427.49 to 438.36, while standard $\text{CLIP}_{\text{B/32}}$ advances from 326.42 to 362.15. Across the Zero-shot MCQ-Neg evaluation, average accuracy ($\textbf{AVG}$) sees substantial gains, such as in $\text{LongCLIP}_{\text{B/16}}$ (35.32 $\rightarrow$ 40.80), $\text{SmartCLIP}_{\text{B/16}}$ (28.37 $\rightarrow$ 38.62), and $\text{NegCLIP}_{\text{B/32}}$ (26.48 $\rightarrow$ 36.01). Notably, across all evaluated backbones, accuracy gains are higher on the \textbf{Hybrid} question subset than on the \textbf{Negated} subset. For instance, in $\text{SmartCLIP}_{\text{B/16}}$, Hybrid performance jumps by $+12.77$, whereas Negated performance increases by only $+2.57$; similarly, base $\text{CLIP}_{\text{B/32}}$ exhibits a $+3.93$ increase on Hybrid (39.46 $\rightarrow$ 43.39) versus $+3.96$ on Negated (6.90 $\rightarrow$ 10.86). While base CLIP models incur a minor trade-off in the \textbf{Affirmed} category, applying our method to semantics-focused architectures enhances Affirmed performance simultaneously ($\text{LongCLIP}_{\text{B/16}}$ rises from 65.06 to 71.85). 

\noindent\textbf{Why Hybrid helps more than pure Negated.} Table \ref{tab:negbench} demonstrates that HANCLIP achieves larger performance gains on Hybrid questions than on Pure Negated ones. This disparity reflects the fundamental difference in question structure: Hybrid queries provide an explicit positive anchor (``A''), enabling HANCLIP's geometric losses to compare inter-clause compatibility and selectively suppress inconsistent features (``not B''). In contrast, Pure Negated queries demand absolute absence reasoning without a supporting visual anchor. Because standard training captions rarely frame missing objects as explicit visual entities, models are prone to relying on superficial lexical cues or bag-of-words biases \cite{yuksekgonul2023negclip}. Furthermore, language models used in template-based caption generation (Table \ref{template}) rarely state an object's absence explicitly, introducing additional upstream bias. Consequently, HANCLIP learns relative multi-clause reasoning far more effectively than absolute absence reasoning, that is consistently observed across all evaluated backbones.

\vspace{-1em}
\subsection{Classification and Retrieval tasks}

\setlength{\textfloatsep}{2pt}
\begin{table*}[!t]
\centering
\small
\resizebox{\linewidth}{!}{
\begin{tabular}{lll|c c c|c c c c | c c c c}
\toprule
\multirow{3}{*}{\textbf{Model}} 
& \multirow{3}{*}{\textbf{ViT}} 
& \multirow{3}{*}{\textbf{Setting}}
& \multicolumn{3}{c|}{\textbf{Classification}}
& \multicolumn{4}{c|}{\textbf{Tradition retrieval}} & \multicolumn{4}{c}{\textbf{Long text retrieval}}\\

\cmidrule(lr){4-6}
\cmidrule(lr){7-10}
\cmidrule(lr){11-14}

& & &
\texttt{ImageNet} & \texttt{CIFAR-10} & \texttt{CIFAR-100}
& \multicolumn{2}{c}{\texttt{COCO}}
& \multicolumn{2}{c|}{\texttt{Flickr30k}}
& \multicolumn{2}{c}{\texttt{Urban1k}} & \multicolumn{2}{c}{\texttt{ShareGPT4V}}\\

\cmidrule(lr){7-8}\cmidrule(lr){9-10}\cmidrule(lr){11-12}\cmidrule(lr){13-14}& & &Acc & Acc & Acc
& I2T & T2I
& I2T & T2I
& I2T & T2I & I2T & T2I\\
\midrule

\multirow{6}{*}{CLIP}
& \multirow{2}{*}{B/16} & Base
& 68.35 & 90.81 & 67.22 & 51.72 & 32.66 & 43.25 & 24.70
& 67.50 & 53.30 & 84.50 & 79.60 \\

& & + Ours 
& 68.02\down & 91.81\up & 69.13\up 
& 53.82\up & 37.56\up & 46.64\up & 30.12\up 
& 69.60\up & 63.70\up & 88.00\up & 85.00\up \\
\cline{2-14}

& \multirow{2}{*}{B/32} & Base
& 63.34 & 89.76 & 64.31 & 50.00 & 30.37 & 39.45 & 21.51
& 61.20 & 46.60 & 83.10 & 77.20 \\

& & + Ours
& 63.00\down & 89.36\down & 65.35\up 
& 52.32\up & 33.86\up & 42.81\up & 25.50\up
& 63.50\up & 56.90\up & 85.30\up & 81.20\up \\
\cline{2-14}

& \multirow{2}{*}{L/14} & Base
& 75.53 & 95.70 & 76.44 & 56.10 & 35.33 & 47.70 & 27.89
& 68.20 & 56.00 & 84.20 & 83.60 \\

& & + Ours
& 75.64\up & 95.71\up & 78.15\up 
& 56.48\up & 37.71\up & 49.52\up & 30.96\up
& 70.40\up & 59.90\up & 87.00\up & 85.50\up \\
\midrule

\multirow{4}{*}{LongCLIP}
& \multirow{2}{*}{B/16} & Base
& 66.82 & 90.76 & 69.16 & 57.30 & 40.36 & 47.19 & 33.17
& 79.30 & 79.40 & 94.80 & 93.30 \\

& & + Ours
& 67.01\up & 90.92\up & 69.67\up 
& 58.04\up & 40.18\down & 50.91\up & 33.57\up
& 81.50\up & 81.10\up & 95.60\up & 93.40\up \\
\cline{2-14}

& \multirow{2}{*}{L/14} & Base
& 72.82 & 96.12 & 79.67 & 62.86 & 46.34 & 53.53 & 41.26
& 82.50 & 86.10 & 97.20 & 97.30 \\

& & + Ours
& 74.13\up & 95.79\down & 79.95\up & 63.78\up & 46.56\up & 57.15\up & 42.20\up
& 85.30\up & 86.80\up & 96.90\down & 97.30 \\
\midrule

\multirow{4}{*}{SmartCLIP}
& \multirow{2}{*}{B/16} & Base
& 66.55\up & 90.69\down & 68.73\down & 61.90 & 42.42 & 55.61 & 36.32
& 90.50 & 87.30 & 98.70 & 98.10 \\

& & + Ours
& 66.11\up & 91.50 & 69.02\up 
& 60.12\down & 42.87\up & 54.83\down & 36.38\up
& 90.70\up & 86.80\down & 98.60\down & 97.70\down \\
\cline{2-14}

& \multirow{2}{*}{L/14} & Base
& 72.51 & 95.81 & 78.29 & 66.08 & 48.44 & 63.98 & 43.84
& 93.30 & 90.10 & 97.90 & 98.50 \\

& & + Ours
& 72.85\up & 96.03\up & 78.23\down 
& 65.70\down & 48.41\down & 64.55\up & 44.02\up
& 93.00\down & 89.50\down & 97.70\down & 98.60\up \\ 
\bottomrule
\end{tabular}}
\caption{Classification and Tradition Retrieval Benchmark. $\textcolor{red}{\uparrow}$ indicates improvement over Base, $\textcolor{blue}{\downarrow}$ indicates degradation.}\vspace{4mm}
\label{tab:classification_and_retrieval}\vspace{-2em}
\end{table*}

\noindent\textbf{Datasets and Metrics.} As same as LongCLIP \cite{zhang2024longclip}, we further evaluate our methodology on the zero-shot image-to-text (I2T) and text-to-image (T2I) retrieval and classification task with the following datasets:
\begin{enumerate}\vspace{-.5em}
    \item \textbf{Classification}: \texttt{CIFAR-10} and \texttt{CIFAR-100} \cite{krizhevsky2009cifar} and \texttt{ImageNet} \cite{russakovsky2015imagenet}.
    \item \textbf{Tradition retrieval} (Short T2I retrieval): \texttt{COCO} \cite{lin2015microsoftcococommonobjects} and \texttt{Flickr30K} \cite{young2014flickr}.
    \item \textbf{Long text-to-image retrieval}: \texttt{Urban1k} \cite{zhang2024longclip} and \texttt{ShareGPT4V} \cite{chen2024sharegpt4v}.
\end{enumerate}\vspace{-.5em}
We use with \textit{Accuracy} as the key metric for Classification task. While COCO and Flickr30k offer the short captions to evaluate the model on coarse-grained retrieval ability, Urban1k and ShareGPT4V were built to evaluate models on long description (average 101 words). We use I2T and T2I Recall at 1 (R@1) as the main metrics \cite{zhang2024longclip}.

\noindent\textbf{Baselines.} We compare every HANCLIP model with its ancestors including CLIP, LongCLIP and SmartCLIP.

\noindent\textbf{Dicussion.} Table \ref{tab:classification_and_retrieval} shows that HANCLIP generally improves image-text retrieval while largely preserving classification performance. The gains are particularly clear for CLIP. For example, in Long text-to-image retrieval, CLIP family improves by up to nearly $+2.52/+5.98$ on I2T/T2I average. Besides, LongCLIP also benefits through almost tasks, and slightly suffers from degradation. Classification changes are modest that the largest gain is $+1.91$ points on CLIP-B/16 CIFAR-100. SmartCLIP shows a clearer trade-off, with several decreases in long-text retrieval, including ShareGPT4V T2I for SmartCLIP-L/14 ($98.50\rightarrow98.60$). Overall, HANCLIP provides strong retrieval improvements, especially for CLIP and LongCLIP, but its transfer benefits remain dependent on the backbone.

\vspace{-1.5em}
\subsection{Impact of Training Objectives} 
\vspace{-2em}
\begin{figure}[!ht]
    \centering
    \begin{minipage}[b]{0.5\linewidth}
        \centering
        \includegraphics[width=\linewidth]{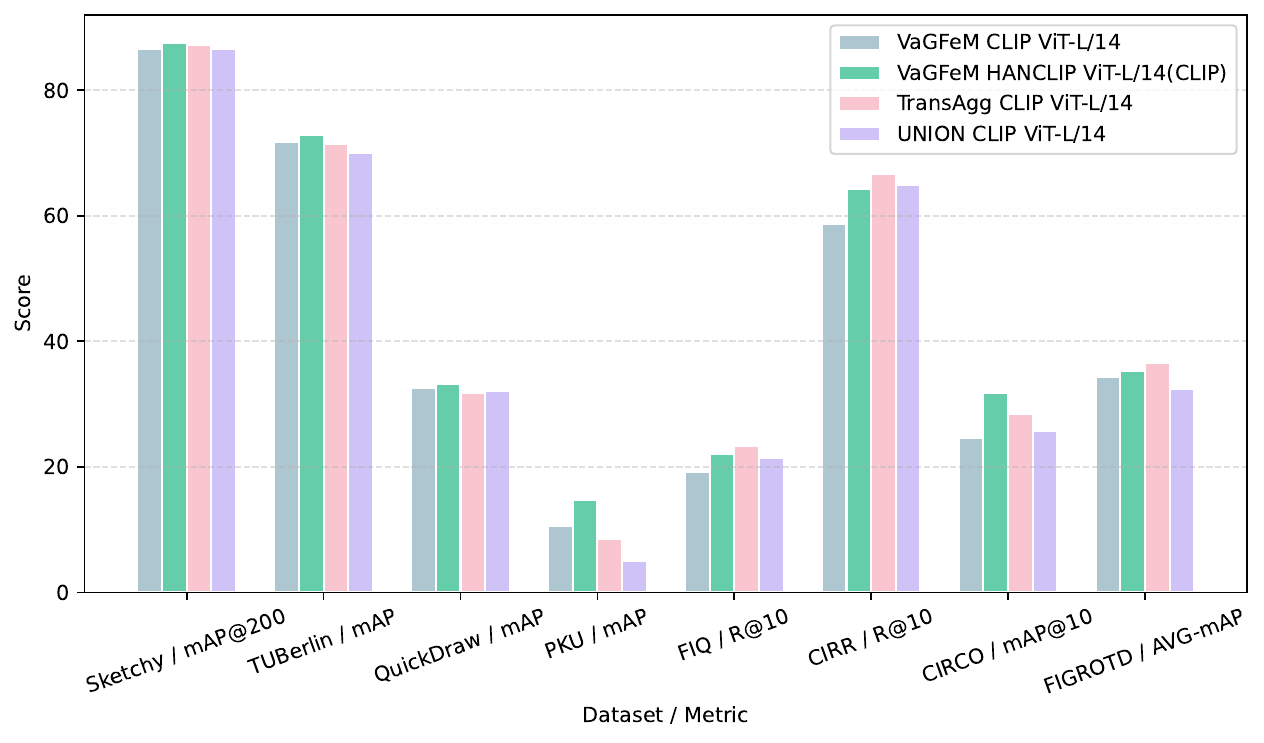}
        \caption{SBIR, CIR and IGROT performance across HANCLIP and CLIP ViT-L/14 based methods.}
        \label{fig:igrot_performance}
    \end{minipage}%
    \hfill
    \begin{minipage}[b]{0.48\linewidth}
        \centering
        \resizebox{\linewidth}{!}{
\begin{tabular}{c cc | ccc cc cc}
\toprule
\multirow{2}{*}{\textbf{Model}} &
\multirow{2}{*}{$\mathcal{L}_{\text{hc}}$} &
\multirow{2}{*}{$\mathcal{L}_{\text{at}}$} &
\multicolumn{3}{c}{\textbf{NegBench}} &
\multicolumn{2}{c}{\textbf{Urban1k}} &
\multicolumn{2}{c}{\textbf{Flickr30k}} \\

\cmidrule(lr){4-6}
\cmidrule(lr){7-8}
\cmidrule(lr){9-10}

& & &
I2T & T2I & MCQ &
I2T & T2I &
I2T & T2I \\
\midrule

\multirow{3}{*}{CLIP}
&  &  &
\cellcolor{pastelred}50.88 & \cellcolor{pastelred}29.06 & 37.99 &
\cellcolor{pastelred}68.20 & \cellcolor{pastelred}56.00 &
\cellcolor{pastelred}47.70 & \cellcolor{pastelred}27.89 \\

& \checkmark &  &
\cellcolor{pastelgreen}53.36 & 32.01 & \cellcolor{pastelred}37.20 &
69.00 & 58.40 &
49.16 & 29.08 \\

& \checkmark & \checkmark &
53.12 & \cellcolor{pastelgreen}34.60 & \cellcolor{pastelgreen}40.06 &
\cellcolor{pastelgreen}70.40 & \cellcolor{pastelgreen}59.90 &
\cellcolor{pastelgreen}49.52 & \cellcolor{pastelgreen}30.96 \\

\midrule

\multirow{3}{*}{LongCLIP}
&  &  &
\cellcolor{pastelred}56.36 & \cellcolor{pastelred}40.95 & \cellcolor{pastelred}37.30 &
\cellcolor{pastelred}82.50 & \cellcolor{pastelred}86.10 &
\cellcolor{pastelred}53.53 & \cellcolor{pastelred}41.26 \\

& \checkmark &  &
60.74 & 43.33 & 38.64 &
84.00 & 86.20 &
57.01 & 41.75 \\

& \checkmark & \checkmark &
\cellcolor{pastelgreen}61.12 & \cellcolor{pastelgreen}43.97 & \cellcolor{pastelgreen}43.86 &
\cellcolor{pastelgreen}85.30 & \cellcolor{pastelgreen}86.80 &
\cellcolor{pastelgreen}57.15 & \cellcolor{pastelgreen}42.20 \\
\midrule

\multirow{3}{*}{SmartCLIP}
&  &  &
\cellcolor{pastelred}63.00 & \cellcolor{pastelred}43.75 & 36.24 &
\cellcolor{pastelgreen}93.30 & \cellcolor{pastelgreen}90.10 &
\cellcolor{pastelred}63.98 & 43.84 \\

& \checkmark &  &
64.02 & 45.39 & \cellcolor{pastelred}35.27 &
\cellcolor{pastelred}92.80 & \cellcolor{pastelred}89.10 &
64.23 & \cellcolor{pastelred}43.80 \\

& \checkmark & \checkmark &
\cellcolor{pastelgreen}64.28 & \cellcolor{pastelgreen}46.29 & \cellcolor{pastelgreen}41.63 & 93.00 & 89.50 &
\cellcolor{pastelgreen}64.55 & \cellcolor{pastelgreen}44.02 \\





\bottomrule
\end{tabular}
}
\vspace{0.6em}
\caption{Comparison using pastel color coding (best \textcolor{pastelgreen}{\rule{0.9em}{0.8ex}}  \textcolor{white}{\rule{0.9em}{0.8ex}} \textcolor{pastelred}{\rule{0.9em}{0.8ex}} worst). $\mathcal{L}_\text{hc}$ and $\mathcal{L}_\text{at}$ represent for hyclip and angular loss. We report results on ViT-L/14-based models.}
\label{tab:loss_ablation}
    \end{minipage}
\end{figure}

\begin{figure}[!ht]
    \centering
    \subfloat[CLIP ViT-B/32]{
        \includegraphics[width=0.45\linewidth]{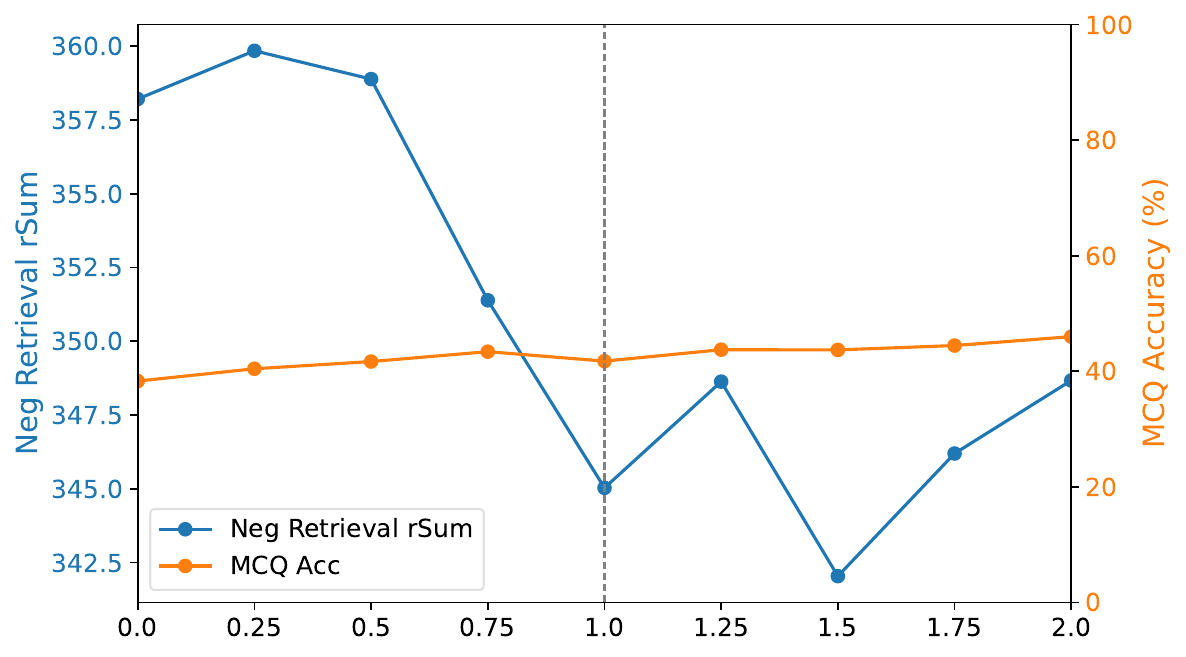}
        \label{fig:vitb32}
    }
    \hfill
    \subfloat[LongCLIP-B]{
        \includegraphics[width=0.45\linewidth]{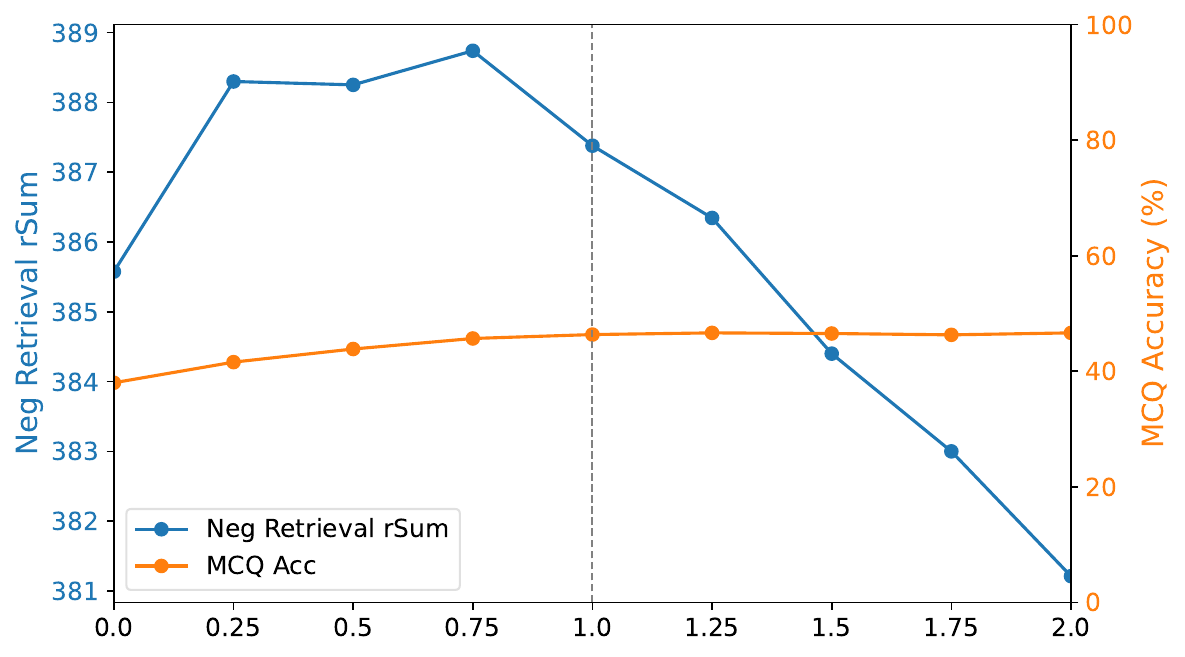}
        \label{fig:longclip_b}
    }

    \vspace{0.5em}

    \subfloat[CLIP ViT-L/14]{
        \includegraphics[width=0.45\linewidth]{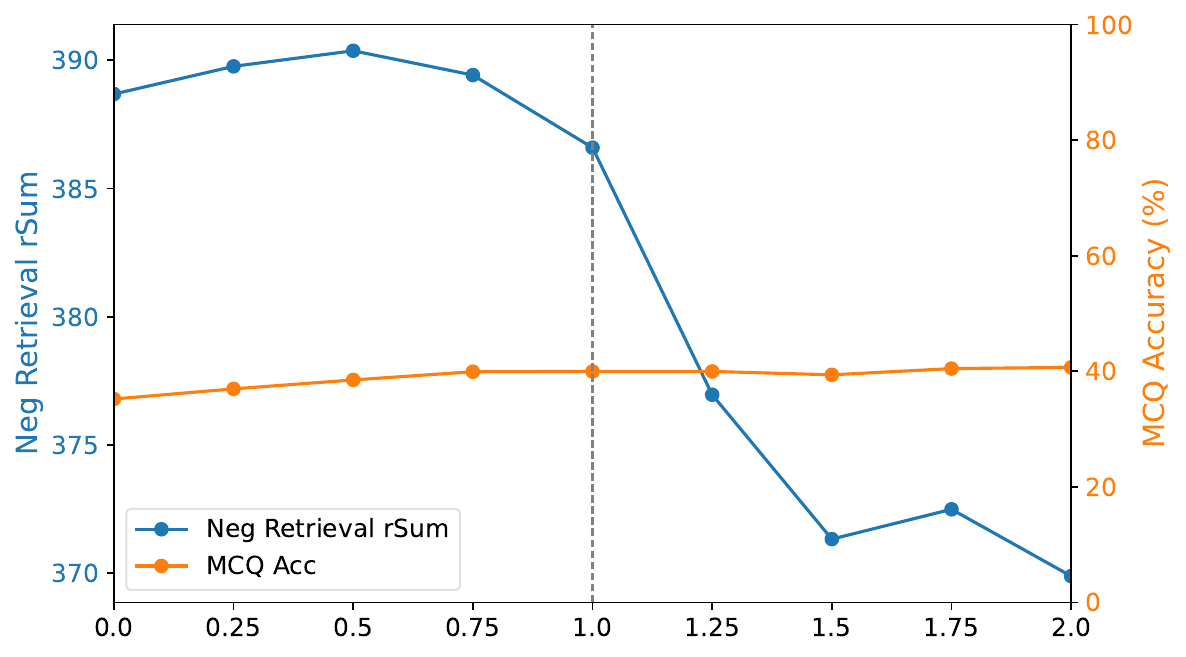}
        \label{fig:vitl14}
    }
    \hfill
    \subfloat[LongCLIP-L]{
        \includegraphics[width=0.45\linewidth]{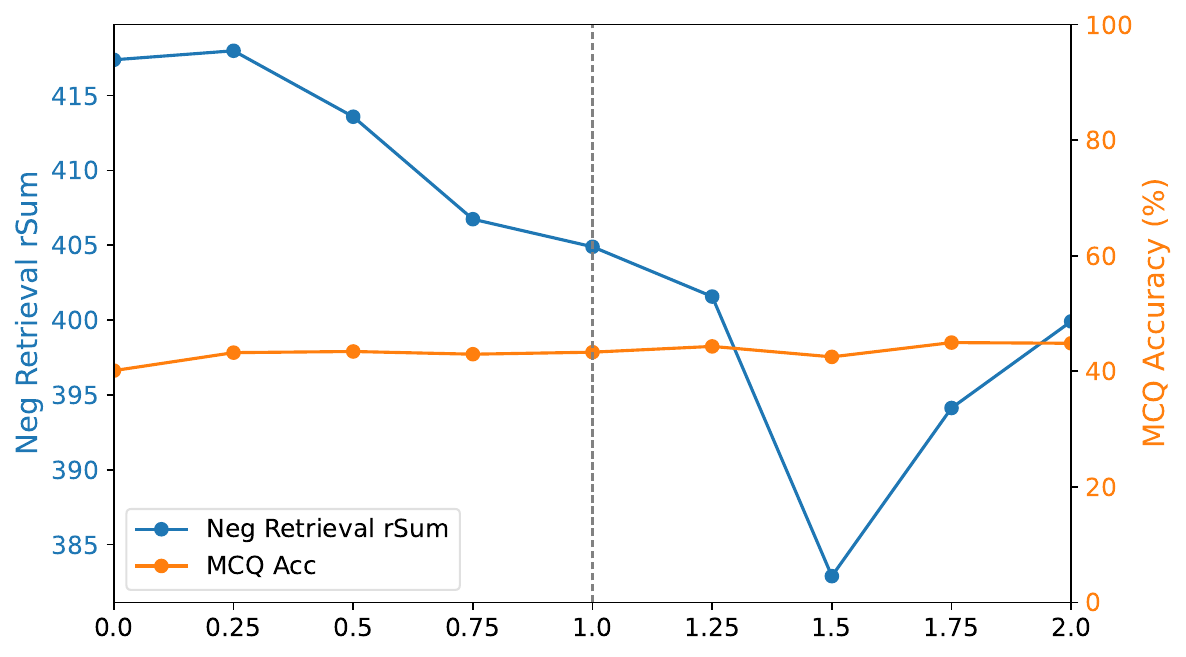}
        \label{fig:longclip_l}
    }

    \vspace{0.5em}
    \caption{Performance of CLIP and LongCLIP on NegBench with different $\alpha$.}
    \label{fig:alpha}\vspace{-2em}
\end{figure}

Table \ref{tab:loss_ablation} show that the hyperbolic contrastive loss
$\mathcal{L}_{\mathrm{hc}}$ and angular loss $\mathcal{L}_{\mathrm{at}}$ are
complementary. Adding only $\mathcal{L}_{\mathrm{hc}}$ improves CLIP's NegBench
I2T/T2I scores from \(50.88/29.06\) to \(53.36/32.01\), while combining both
objectives further changes them to \(53.12/34.60\) and raises MCQ accuracy
from \(37.99\) to \(40.06\). The same trend holds for LongCLIP and SmartCLIP:
the combined objective achieves the best NegBench MCQ scores of \(43.86\) and
\(41.63\), respectively. It also provides consistent gains on Urban1k and
Flickr30k retrieval, demonstrating that the angular objective complements
hyperbolic contrastive learning by improving negation discrimination without
sacrificing general image-text alignment.

Additionally, figure~\ref{fig:alpha} reveals a clear trade-off between negated retrieval (rSum) and MCQ accuracy across backbones. Negated retrieval improves up to $\alpha \in [0.5, 0.75]$, but drops sharply beyond $\alpha = 1.0$ as heavy negation weighting distorts the ranking geometry. Conversely, MCQ accuracy increases smoothly and saturates, benefiting from the altered space even when ranking suffers. Setting $\alpha = 1.0$ provides the optimal balance, maximizing retrieval rSum while retaining near-peak MCQ gains across all models.
\vspace{-1em}
\subsection{Performance On Other Retrieval Tasks}

Image-Guided Retrieval with Optional Text (IGROT) \cite{agnolucci2024isearle} unifies Composed Image Retrieval (CIR) \cite{agnolucci2024isearle} and Sketch-Based Image Retrieval (SBIR) \cite{liu2017sketchy} by querying with an anchor image optionally modified by text.

\noindent\textbf{Datasets and Metrics.} We train using 5,000 samples from LlavaSCo (CIR) and Training-\texttt{Sketchy} \cite{le2025union} (SBIR), and validate on \texttt{FIGROTD} \cite{le2025figrotd}. We evaluate SBIR on \texttt{Sketchy}~\cite{yelamarthi2018zssbir}, \texttt{TUBerlin}~\cite{zhang2016tuberlin}, \texttt{QuickDraw}~\cite{dey2019quickdraw}, and \texttt{PKU-Sketch}~\cite{pang2018pku} (reporting mAP). For CIR, we evaluate on \texttt{FIQ}~\cite{guo2019fashion} and \texttt{CIRR} (R@10), and \texttt{CIRCO}~\cite{agnolucci2024isearle} (mAP@10). Finally, we report AVG-mAP on the \texttt{FIGROTD} benchmark, which spans SBIR, CIR, and Composed Sketch Text-Based Image Retrieval (CSTBIR).

\noindent\textbf{Baselines.} We evaluate VaGFeM \cite{le2025figrotd} using both standard CLIP ViT-L/14 and our HANCLIP backbone. We compare against TransAgg \cite{liu2023zeroshot} and UNION \cite{le2025union}.

\noindent\textbf{Discussion.} Figure \ref{fig:igrot_performance} shows HANCLIP consistently enhances retrieval across tasks. In SBIR, it notably improves on abstract sketches (\texttt{QuickDraw}, \texttt{PKU}). For CIR, HANCLIP outperforms standard CLIP by $+4$ R@10 on \texttt{CIRR} and $+5$ mAP@10 on \texttt{CIRCO}, showing superior compositional reasoning, though it trails TransAgg slightly ($2$ mAP@10 on \texttt{CIRCO}). On \texttt{FIGROTD}, HANCLIP lifts AVG-mAP from 34.26 to 35.50 ($+3.6$). Overall, HANCLIP delivers robust, cross-task improvements, balancing generality and performance against highly specialised aggregation methods.
\section{Related Works and Discussions}
\noindent\textbf{Vision-Language Models.} Since the introduction of Vision Transformers (ViT) \cite{dosovitskiy2020vit}, foundational vision-language models (VLMs) such as CLIP \cite{radford2021learning}, BLIP \cite{li2022blip,li2023blip2}, and ALIGN \cite{jia2021align} have relied primarily on contrastive pre-training objectives \cite{gao2021simcse}. To improve representation quality and efficiency, subsequent works refine this objective through loss innovations—such as SigLIP's sigmoid loss \cite{zhai2023sigmoid,tschannen2025siglip2}, CyCLIP \cite{shashank2022cyclip}, RankCLIP \cite{zhang2025rankclip}, AlignCLIP \cite{eslami2025alignclip}, and region-attentive objectives like AlphaCLIP \cite{zeyi2024alphaclip}. Beyond general pre-training, VLMs have also been tailored for specialized tasks, including Composed Image Retrieval (MagicLens \cite{zhang2024magiclens}, TransAgg \cite{liu2023transagg}, UNION \cite{le2025union}, FIGROTD \cite{le2025figrotd}) and Object Detection (e.g., the DetCLIP series \cite{yao2022detclip,yao2023detclipv2,pi2024insdetclip,yao2024detclipv3}). In contrast to conventional contrastive functions, we propose a new learning objective built on CLIP \cite{radford2021learning} that incorporates explicit negative group sampling to improve semantic discrimination.

\noindent\textbf{Negation Alignment in VLMs.}
Yuksekgonul \etal~\cite{yuksekgonul2023negclip} first observed that vision-language models (VLMs) often reduce text-image relationships to simple bag-of-words semantics, proposing NegCLIP to augment training with synthetic hard negatives. Subsequent works fine-tune VLMs on specialized negation data: CoNCLIP~\cite{singh2025conclip} leverages negated pairs from CC-Neg, NegationCLIP~\cite{park2025negationclip} uses MLLM-generated negation captions, and CECLIP~\cite{le2024ceclip} applies intra- and cross-modal hard negative ranking. Alternatively, NEAT~\cite{han2025neat} introduces a test-time adaptation framework using unlabeled negation data, while NegBench~\cite{alhamoud2025negbench} establishes a benchmark across 18 task variations to systematically evaluate negation handling. However, these methods require extensive training data. In contrast, we extract a concise 20,000-quadruplet subset from CC-Neg~\cite{singh2025conclip} and introduce the Angular Triplet Negation loss. This objective regulates distances between opposing semantic groups without degrading general representation quality.

\noindent\textbf{Non-Euclidean Vision Language Models.}
Hyperbolic Vision Transformers \cite{ermolov2022hyperbolic} pioneered non-Euclidean vision backbones, demonstrating superior semantic hierarchy and uncertainty modeling over Euclidean baselines. MERU \cite{desai2023meru} extended this to multimodal learning via Lorentzian image-text embeddings that capture concept entailment. Subsequent works further expanded hyperbolic VLM capabilities: HyCoCLIP \cite{pal2024hycoclip} incorporated multi-granularity compositional lattices across images, boxes, and text; HyperVLM \cite{srivastava2025hypervlm} refined Poincaré guidance for coarse-to-fine taxonomies; and PHyCLIP \cite{yoshikawa2025phyclip} leveraged an $\ell_1$-product of hyperbolic spaces to model complex cross-family hierarchies. Unlike prior works focused primarily on structural hierarchy, our hyperbolic formulation equips VLMs with robust negation understanding while preserving performance on traditional retrieval and classification tasks.

\vspace{-1em}
\section{Conclusion} \vspace{-.5em}
In this study, we investigate why Vision--Language Models struggle with fine-grained negation despite strong contrastive objectives. We find that standard CLIP-style losses treat negated and affirmative captions coarsely, failing to establish a distinct representational structure for absent concepts. To address this, we propose a negation-aware framework that combines hyperbolic geometry with an angle-based training objective to explicitly organize affirmed, negated, and hybrid descriptions into structurally separate regions of the embedding space. Evaluated across four VLMs, our geometry-aware approach consistently improves negation sensitivity and multiple-choice reasoning while preserving strong retrieval performance, demonstrating that directly embedding logical structure into feature geometry provides a principled, scalable path to advance VLM reasoning without large-scale retraining.
\bibliographystyle{splncs04}
\bibliography{myref}
\end{document}